\pdfoutput=1

\documentclass[11pt]{article}

\usepackage[final]{acl}

\usepackage{times}
\usepackage{hyperref}
\usepackage{latexsym}
\usepackage{longtable}
\usepackage{multirow}
\usepackage{array}
\usepackage{booktabs}
\usepackage{enumitem}
\usepackage{amsmath}
\usepackage[T1]{fontenc}

\usepackage[utf8]{inputenc}

\usepackage{microtype}

\usepackage{inconsolata}

\usepackage{graphicx}

\usepackage{xcolor}
\title{\textsc{Heart}-felt Narratives: \\Tracing Empathy and Narrative Style in Personal Stories with LLMs}

\author{
    Jocelyn Shen$^1$\enspace\quad
    Joel Mire$^{2}$ \quad
    \textbf{Hae Won Park$^1$} \\ \quad
    \textbf{Cynthia Breazeal$^1$}\quad \textbf{Maarten Sap$^{2}$ }\\\
    \small{$^1$Massachusetts Institute of Technology, Cambridge, MA, USA} \vspace{-.2em}\\
    \small{$^2$Carnegie Mellon University, Pittsburgh, PA, USA}  \vspace{-.2em}\\
    \normalsize{\texttt{joceshen@mit.edu, jmire@andrew.cmu.edu, haewon@mit.edu, }}\\\normalsize{\texttt{ breazeal@mit.edu, msap2@andrew.cmu.edu}}
}

\begin{document}

\maketitle

\begin{abstract}

  Empathy serves as a cornerstone in enabling prosocial behaviors, and can be evoked through sharing of personal experiences in stories. While empathy is influenced by narrative content, intuitively, people respond to the \textit{way} a story is told as well, through narrative style. Yet the relationship between empathy and narrative style is not fully understood. In this work, we empirically examine and quantify this relationship between style and empathy using LLMs and large-scale crowdsourcing studies. We introduce a novel, theory-based taxonomy, \textsc{Heart} (Human Empathy and Narrative Taxonomy) that delineates elements of narrative style that can lead to empathy with the narrator of a story. We establish the performance of LLMs in extracting narrative elements from \textsc{Heart}, showing that prompting with our taxonomy leads to reasonable, human-level annotations beyond what prior lexicon-based methods can do. To show empirical use of our taxonomy, we collect a dataset of empathy judgments of stories via a large-scale crowdsourcing study with $N=2,624$ participants.\footnote{We make all our annotations, study data results, and language model results publicly available at \url{https://github.com/mitmedialab/heartfelt-narratives-emnlp}} We show that narrative elements extracted via LLMs, in particular, vividness of emotions and plot volume, can elucidate the pathways by which narrative style cultivates empathy towards personal stories. Our work suggests that such models can be used for narrative analyses that lead to human-centered social and behavioral insights. 
\end{abstract}
\begin{figure}[t]
    \centering
    \includegraphics[width=\linewidth]{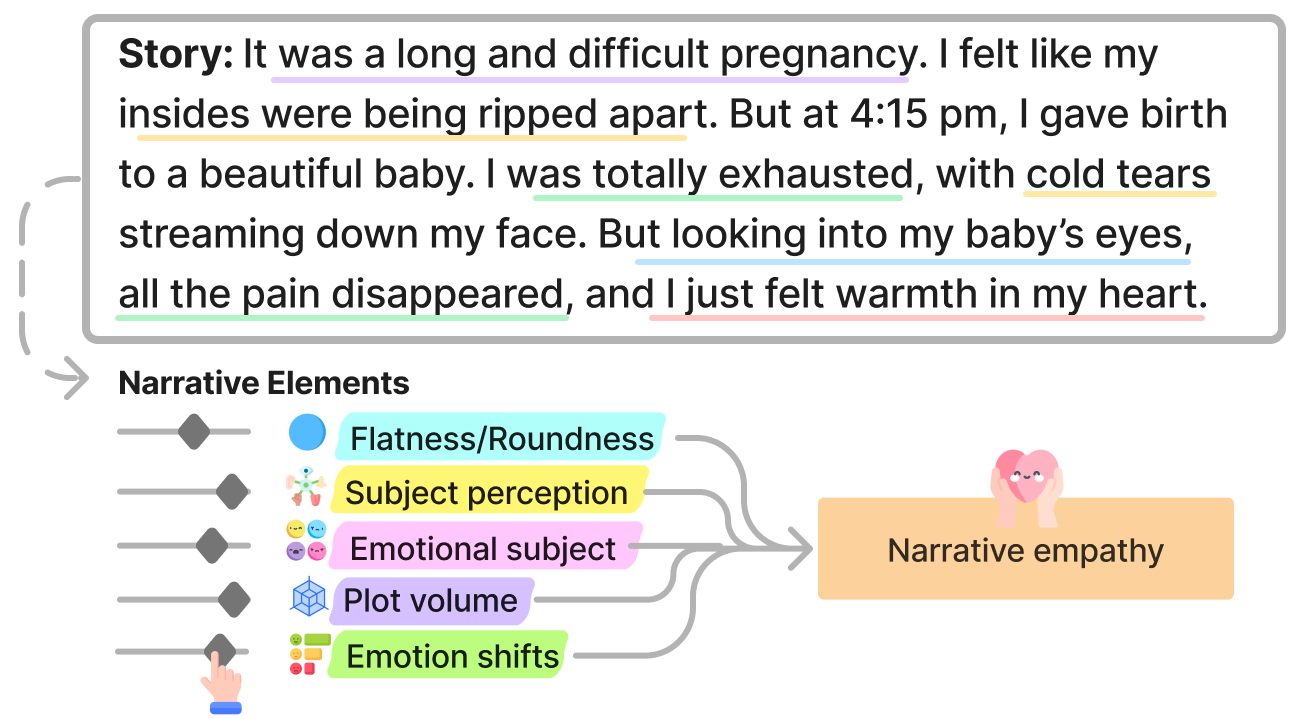}
    \caption{Narrative empathy can be evoked through the \textit{way} a story is told (narrative style). This work introduces \textsc{Heart}, a theory-driven taxonomy of narrative elements that contribute to empathy.
    }
    
    \label{fig:teaser}
    \vspace{-10pt}
\end{figure}

\section{Introduction}
Empathy, which is a foundational psychological process that drives many prosocial functions, \cite{zaki_war_2019, morelli_emerging_2015}, is often delivered through storytelling and sharing of personal experiences \cite{coplan_empathic_2004, keen_narrative_2014}. 
Empathetic responses evoked by stories are affected by factors beyond the content of the story alone -- delivery, context, and reader characteristics all contribute to the emotional resonance of a narrative. Most studies of narrative empathy and its related constructs focus on reader characteristics,and content of a story \cite{sharma_computational_2020, shen_modeling_2023-1}. However, intuitively, people also respond to the \textit{way} a story is told, or the stylistic devices used within a narrative (Figure \ref{fig:teaser}).\footnote{Note that our definition of narrative style may differ slightly from pure traditional stylistics. Aspects of style are naturally intertwined with the content of a story, but our taxonomy focuses more on the ways in which certain content are expressed (for example, rather than focusing on ``what'' emotion is present in the story, targeting instead the vividness of emotional language).} 

A key challenge in narrative analysis within the NLP community is that extracting stylistic features relevant to empathy is not trivial.
Prior works use word-count-based \cite[e.g., lexica;][]{roshanaei_paths_2019, zhou_language_2021} or hand-crafted features on extremely limited story sets \cite{kuzmicova_literature_2017, fernandez-quintanilla_textual_2020, fernandez-quintanilla_introduction_2023, eekhof_engagement_2023, mangen_empathy_2018,hartung_taking_2016}  to quantify narrative elements. However, more complex stylistic narrative devices, such as plot shifts \cite{nabi_role_2015} or vividness of emotions \cite{pillemer_remembering_1992} are harder to summarize with lexica alone. While a few works have explored using LLMs for more complex narrative analysis tasks \cite{zhu_are_2023, michelmann_large_2023, sap_quantifying_2022}, to what extent LLMs can effectively model stylistic devices, and how LLM-extracted features might be leveraged for downstream social insights, remains underexplored.

In this work, we fill this gap by presenting the following contributions.
(1) We introduce \textsc{Heart} (Human Empathy and Narrative Taxonomy), a theory-driven taxonomy of narrative style elements that relate to empathy. (2) We use LLMs to quantify aspects of narrativity in our taxonomy and evaluate how well LLMs represent these elements in line with human judgments. For a subset of narrative elements with available lexica, we compare lexical measures with LLM measures, finding that in most cases, GPT-4 and Llama 3 outperform lexica. (3) Through a human study of $N=2,624$ participants, we introduce a new crowdsourced dataset (\textsc{Heart}-felt Stories Dataset) of empathetic reactions to personal narratives, including annotated narrative style elements, reader characteristics and narrative reactions. (4) With our dataset, we conduct an analysis of pathways through narrative style and reader characteristics leading to empathy, demonstrating the value of \textsc{Heart} in exploring empirical behavioral insights around narrative empathy. In particular, we find that narrative styles with heightened vividness of emotions, character development and action, and plot volume, are tied to narrative empathy. We additionally show that empathy is personalized, with high variability even for the same story, and that beyond narrative style, factors like a reader's trait empathy and similarity of experiences to the narrator also significantly impact empathy. 

\section{Related Work} 
Computational linguistic methods can be used to analyze many aspects of narrativity across a large corpus of stories \cite{sap_quantifying_2022}. Prior works have used lexicon-based approaches to extract psychologically-grounded word categories and relate these to empathy \cite{roshanaei_paths_2019, xiao_computational_2016-1}. \citet{zhou_language_2021} use linguistic style features such as degree of interdependent thinking and integrative complexity (the ability of a person to recognize multiple perspectives and connect them) to predict a viewer's empathy towards a specific situation. 
\citet{antoniak_narrative_2019} apply narrative analysis techniques to birth stories online, and show patterns of affective and event-based sequences over time.
More recently, \citet{yaden_characterizing_2024} used linguistic features, such as word phrases and topics, and leveraged LDA to analyze language that separates more empathetic people from more compassionate people, showing that compassionate people use more other-focused language than empathetic people. Other works leverage recent natural language processing (NLP) methods to predict empathy and prosociality from text \cite{shen_modeling_2023-1, buechel_modeling_2018-1, sharma_computational_2020, bao_conversations_2021}, but do not explore pathways via which readers feel empathy.

A few works have explored the power of LLMs in characterizing aspects of narrative. In particular, \citet{michelmann_large_2023} show that LLMs serve as good approximations of human annotators in narrative event segmentation. Other works show that LLMs achieve reasonable performance on character profiling tasks for fictional narratives, particularly in factual consistency and motivation understanding. However, \citet{subbiah_reading_2024}  indicate that LLMs fail to perform authentic summarization of stories in line with feedback from writers, apart from successfully drawing on thematic components of the stories. Ultimately, LLMs demonstrate growing potential for narrative understanding tasks \cite{zhu_are_2023}, but how well they perform, what types of tasks they succeed in, and how they can reveal human behavioral insights, is an active area of research \cite{agnew_illusion_2024-1}.

Our work leverages LLMs to extract narrative style elements that may play a role in narrative empathy through our grounded taxonomy. We evaluate the performance of prompting LLMs to extract such elements against expert human raters. Our empirical study using LLM-extracted narrative elements focuses more on the scientific and behavioral question of how to untangle aspects of narrative style and reader characteristics to understand their contribution towards empathy, rather than improving performance on empathy prediction alone. 

\section{Background}
Empathy in the context of narratives has been the subject of many studies in psychology and literary studies. We briefly summarize those below. 

\paragraph{Narrative Style and its Role in Empathy.} Prior works have theorized how shifts in narrative style impact empathic effect of a story. \citet{keen_theory_2006} proposed a theory of narrative empathy that draws on narrative techniques to enhance empathy, such as flatness or roundness of a character, the character's mode of consciousness, and vivid use of settings. \citet{van_krieken_evoking_2017} presented a framework of linguistic cues to measure identification with narrative characters, including character dimensions such as the emotional or perceptual subject of the story. This framework covers both background elements of a story, which can facilitate immersive experiences, and foregrounded elements (such as figurative language), which facilitate aesthetic experiences with the text \cite{jacobs_neurocognitive_2015}. 

However, many of these narrative techniques, particularly those that are more abstract in nature, such as plot structure or emotional shifts \cite{nabi_role_2015}, have yet to be tested empirically. Researchers in narratology have explored the impact of literary quality on reader empathy, varying aspects such as foregrounding, point of view/viewpoint words, emotion and discourse presentation, and characterisation techniques, but have found mixed results in small-scale studies  \cite{kuzmicova_literature_2017, fernandez-quintanilla_textual_2020, fernandez-quintanilla_introduction_2023, eekhof_engagement_2023, mangen_empathy_2018,hartung_taking_2016}. Other studies have looked at how aspects of literary reading contribute to transportation, or the ability to absorb in a narrative, which further predicts empathy towards a story \cite{walkington_impact_2020, van_laer_extended_2014, van_laer_what_2019}. \citet{koopman_empathic_2015} conducted a larger-scale study to investigate the role of genre, personal factors, and affective responses on both empathic understanding and pro-social behavior, finding that genre affected prosocial behaviors. However, narrative style encompasses many aspects beyond genre alone, and each of these elements couples with one another to enhance or diminish narrative empathy.

\paragraph{Reader Characteristics and Narrative Empathy.}
While narrative style can have an effect on empathy, other factors such as the reader's characteristics or experiences during reading can affect empathy as well. For example, psychology, economics, and neuroscience have suggested that gender has a significant influence on people's cognitive empathy, with women exhibiting higher cognitive empathy than men across a variety of age groups~\cite{christov2014empathy,michalska2013age, o2013empathic}. Levels of narrative empathy can also be modulated by one's trait empathy level \cite{konrath_development_2018}, emotional state during reading \cite{roshanaei_paths_2019}, or general exposure to literature \cite{mar_bookworms_2006}. Untangling the effects of these fixed can be challenging, and has been attempted by a few prior works, but with varied results \cite{koopman_effects_2015, fernandez-quintanilla_textual_2020, roshanaei_paths_2019}.

In our work, we propose a taxonomy of narrative empathy based on theories and empirical results presented in the aforementioned works, then scientifically explore what pathways through both narrative style and reader characteristics and life experiences and to overall empathy towards a story. In contrast to prior works, which often vary a single element of narrative style, we construct a thorough taxonomy of narrative elements related to empathy. 

\begin{figure*}[t]
    \centering
    \includegraphics[width=.75\textwidth]{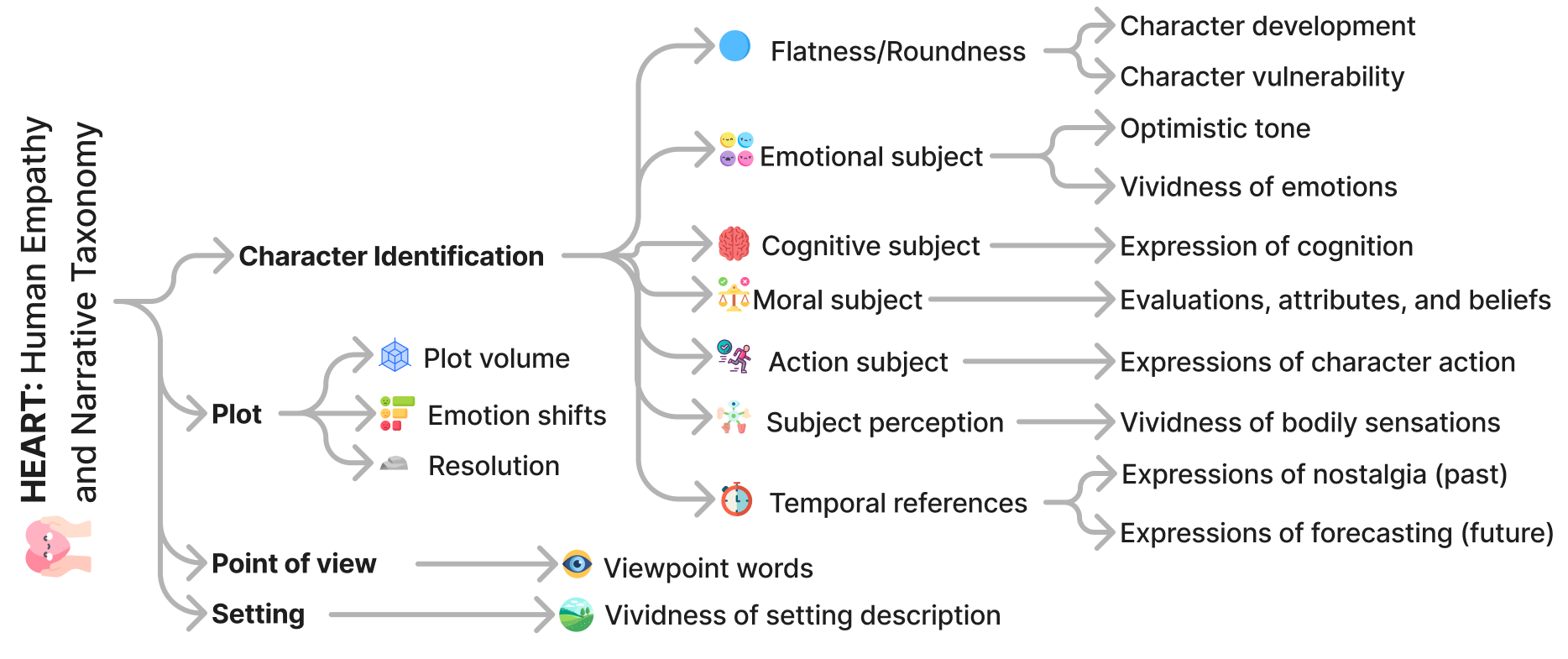}
    \caption{Narrative Empathy and Style Taxonomy delineating aspects of narrative style that theoretically relate to empathy towards a narrative.
    }
    \label{fig:heart}
    \vspace{-5pt}
\end{figure*}

\section{\textsc{Heart} Taxonomy for Empathy and Narrative Style}
\label{sec:heart}

Based on the aforementioned theoretical and empirical research, we propose \textsc{Heart}, a taxonomy of narrative style elements that can lead to empathy. In \textit{A Theory of Narrative Empathy}, Keen posits that aspects of characterization, narrative situation, internal perspective, and techniques to represent character consciousness can contribute to narrative empathy. We use these concepts as precursors for developing \textsc{Heart}. Our theoretical model serves as a starting point for understanding \textit{what} aspects of narrative characteristics might lead to empathy and \textit{how} we can measure these factors using  computational approaches. 

Figure \ref{fig:heart} shows our full taxonomy, which delineates narrative style as it relates to narrative empathy via four main categories: (1) Character identification (2) Plot (3) Point of view and (4) Setting. In the remainder of this section, we outline each element of our taxonomy and the theoretical and empirical roots of how each element may contribute to narrative empathy.

\paragraph{Character Identification} We refer to character identification elements as story aspects that draw readers into the narrator's perspective, whether this be across internal dimensions (emotion/cognition) or external dimensions (perception/time). We define 6 high-level elements of our taxonomy that can contribute to identification with a character in a story, primarily rooted in \cite{van_krieken_evoking_2017}'s work on character identification: 
\begin{enumerate}[itemsep=0em,leftmargin=1.25em,topsep=0.5em]
    \item \textbf{Flatness/roundness} \cite{keen_theory_2006} of the character, including depth of the character expressed through {character development} over the course of the story or {character vulnerability}.
    \item \textbf{Emotional subject} \cite{van_krieken_evoking_2017, roshanaei_paths_2019, pillemer_remembering_1992}, refers to the way emotions are expressed both in tone and {vividness of emotions}.
    \item \textbf{Cognitive subject} \cite{schweitzer_language_2021, van_krieken_evoking_2017}, captures expressions of {cognition} such as thinking, planning, and decision making.
    \item \textbf{Moral subject} \cite{van_krieken_evoking_2017, saldias_exploring_2020} primarily refers to how {evaluations} or expressions of the narrator's opinion are conveyed through the story.
    \item \textbf{Action subject} \cite{van_krieken_evoking_2017}, refers to expressions of character action.
    \item \textbf{Subject perception} \cite{van_krieken_evoking_2017} captures the vividness of {perception and bodily sensations} experienced by the character.
    \item \textbf{Temporal references} \cite{pillemer_remembering_1992} contain expressed nostalgia (looking to the past) or forecasting and anticipation (looking to the future).
\end{enumerate}

\paragraph{Plot} Defining plot has been a key task in narrative analysis \cite{toubia_how_2021, reagan_emotional_2016}, and can foster empathy through enhancing the narrator's story via shifts at critical junctures. We delineate 3 aspects of plot that relate to narrative empathy:
\begin{enumerate}[itemsep=0em, leftmargin=1.25em,topsep=0.5em]
    \item \textbf{Plot volume} \cite{keen_narrative_2014, van_laer_extended_2014, van_laer_what_2019} captures the frequency and significance of events in a story. 
    \item \textbf{Emotion shifts} \cite{nabi_role_2015} indicate fluctuations in the overall emotional trajectory of the story (such as from low to high valence and vice versa). 
    \item \textbf{Resolution} \cite{mcadams_problem_2006} captures the release of tension after the main conflict that a character experiences.
\end{enumerate}

\paragraph{Point of view} Prior works suggest that point of view can affect empathy towards a narrator  \cite{eekhof_engagement_2023, fernandez-quintanilla_textual_2020, spitale_socially_2022}. For example, first-person perspective can emphasize the personal nature of the story and draw readers into the shoes of the narrator.

\paragraph{Setting} Finally, the environment and context of the narrator can facilitate narrative empathy \cite{pillemer_remembering_1992, van_krieken_evoking_2017}, for example through world-building to enhance narrative transportation. We capture this element via the vividness of the setting description in a narrative. 

\section{\textsc{Heart}-felt Stories Dataset Annotation}
With our theory-grounded taxonomy, we next evaluate how well LLMs can approximate narrative style elements. In order to do so, we annotate the \textsc{Heart}-felt Stories Dataset, a corpus of personal narratives with expert ratings on a subset of stories.
\subsection{Story Dataset}
To empirically observe the narrative elements of \textsc{Heart}, we started with a seed dataset of personal narratives from the \textsc{ EmpathicStories }\cite{shen_modeling_2023-1} and the \textsc{EmpathicStories++} \cite{shen2024empathicstories} dataset, which were specifically designed to include meaningful and vulnerable personal stories with diverse narrators, shared across diverse topics (e.g. relationships, mental health, career and school, etc.). The \textsc{EmpathicStories} dataset consists of $\sim$1,500 personal narratives collected from social media sites (Facebook, Reddit), crowdsourced personal narratives, and transcribed podcasts. The \textsc{EmpathicStories++} dataset contains $ \sim$500 conversational personal stories that were automatically transcribed from storytelling interactions with an AI. 
We filtered stories to remove potentially harmful topics (e.g. mentions of sexual assault, excessive swearing), and filtered stories that were under 200 words (which might not contain rich narrative style elements), resulting in a final dataset of 874 personal stories.

\begin{table}[t]

    \centering
        
    \footnotesize
    \resizebox{\linewidth}{!}{
    \begin{tabular}{rrrl} 
\textbf{Feature} & \textbf{KA} & \textbf{PPA} & \textbf{$\rho$} \\
\hline Optimistic tone & 49.27 & 81.50 & $72.21^{* * *}$ \\
Vivid setting & 48.48 & 76.00 & $64.23^{* * *}$ \\
Plot volume & 45.97 & 83.50 & $60.32^{* * *}$ \\
Resolution & 44.29 & 79.00 & $58.97^{* * *}$ \\
Character vulnerability & 38.17 & 75.00 & $50.06^{* * *}$ \\
Character development & 28.55 & 72.50 & $45.24^{* *}$ \\
Cognition & 27.56 & 70.00 & $39.18^{* *}$ \\
Evaluations & 26.29 & 74.00 & $31.3^*$ \\
Emotion shifts & 23.49 & 74.50 & $46.34^{* *}$ \\
Vivid emotions & 21.17 & 66.00 & $31.8^*$ \\
Temporal references & 18.29 & 77.00 & $27.96^*$ \\
Bodily sensations & 3.79 & 60.33 & $34.25^*$
\end{tabular}}
\caption{Agreement between 2 expert human annotators on the narrative elements of our taxonomy. Scores are multiplied by 100 and rounded for readability and sorted by KA. Spearman's correlation $\rho$ indicates significance. 
}
\vspace{-5pt}
    \label{tab:humanagreement}
\end{table}

\subsection{Expert Narrative Style Annotation}\label{ssec:expert-annotations}
We randomly sampled 50 stories from our final dataset of 874 stories to obtain expert annotations of the narrative elements and validate LLM performance on the task. We selected a subset of 12 narrative elements from our taxonomy that are non-trivial to extract from existing NLP toolkits, and which required human judgments given the subjectivity of the task.
Three independent members of our research team with expertise in text analysis and annotation iteratively designed a codebook (Appendix \ref{codebook}) with instructions and examples for gauging the presence of each element.

Subsequently, two independent expert annotators rated the presence of each of the 12 narrative elements in the 50 sampled stories. Table \ref{tab:humanagreement} shows the agreement between the 2 raters using Krippendorf's alpha (KA), percent pairwise agreement (PPA), and Spearman's correlation ($\rho$). All ratings are positively correlated to each other, but different narrative elements have varying degrees of agreement. We observe the lowest agreement between human annotators for \textsc{temporal references} and \textsc{bodily sensations}, where irrealis events and mentions of body sensations across multiple characters caused confusion. Moreover, while some human agreements may appear low using the KA metric, these scores are consistent with prior NLP tasks with more subjectivity \cite{shen_modeling_2023-1, rashkin_modeling_2018, sap_connotation_nodate}. In our subsequent empirical analysis, we do not use features with low agreement (below $0.2$ KA). 


\begin{table*}[t]
    \centering
    \footnotesize
    \resizebox{0.7\textwidth}{!}{
    \begin{tabular}{r|ccc|ccc} 
    & \multicolumn{3}{c|}{\textbf{GPT-4}} & \multicolumn{3}{c}{\textbf{Llama 3 8B Instruct}} \\
    \cline{2-7}
    \textbf{Feature} & \textbf{KA} & \textbf{PPA} & \textbf{$\rho$} & \textbf{KA} & \textbf{PPA} & \textbf{$\rho$} \\
    \hline
    Character vulnerability & 62.89 & 86.50 & $80.15^{***}$ & 27.08 & 79.00 & 70.55*** \\
    Optimistic tone & 50.97 & 82.25 & $68.06^{***}$ & 48.41 & 82.75 & 67.14*** \\
    Resolution & 44.55 & 80.00 & $61.59^{***}$ & 7.26 & 71.83 & 34.93* \\
    Character development & 44.09 & 79.25 & $61.64^{***}$ & 20.99 & 77.25 & 46.51** \\
    Vivid setting & 42.12 & 78.00 & $67.31^{***}$ & -31.07 & 57.03 & 41.84** \\
    Plot volume & 33.00 & 79.25 & $44.88^{**}$ & -4.00 & 76.08 & 27.51 \\
    Emotion shifts & 32.25 & 82.25 & $45.5^{**}$ & 25.13 & 80.58 & 52.13*** \\
    Vivid emotions & 27.25 & 75.00 & $59.21^{***}$ & 25.80 & 76.00 & 42.13** \\
    Cognition & 19.83 & 73.00 & $34.91^{*}$ & 24.98 & 76.00 & 52.89*** \\
    Evaluations & -9.76 & 75.00 & 22.69 & -27.16 & 73.00 & \textit{NaN} \\

    \end{tabular}
    }
    \caption{Agreement between aggregated human annotators (gold ratings) and GPT-4 and Llama 3 8B Instruct ratings of narrative elements in our taxonomy. 
    Rows are sorted by GPT-4 KA. 
    }
    \label{tab:gptagreement}
    \vspace{-10pt}
\end{table*}





\begin{table}[t]
    \centering
     
    \footnotesize
    \resizebox{1.0\linewidth}{!}{
   \begin{tabular}{rlll} 
\textbf{Feature} & \textbf{$\rho_{LIWC}$} & \textbf{$\rho_{GPT-4}$} & \textbf{$\rho_{Llama3}$} \\
\hline
Optimistic tone& $47.35^{* * *}$ & \textbf{$68.06^{* * *}$} & \textbf{$67.14^{* * *}$} \\
Cognition & \textbf{$41.29^{* *}$ }& $34.91^*$ & \textbf{$52.89^{* * *}$} \\
Vivid emotions & $37.63^{* *}$ & \textbf{$59.21^{* * *}$} & $42.13^{* *}$ \\
Character vulnerability & -6.95 &\textbf{$80.15^{* * *}$} & \textbf{$70.55^{* * *}$}
\end{tabular}
}
   \caption{Comparison of correlations with human annotations for LIWC, GPT-4, and Llama 3 8B Instruct.
        }
            \label{tab:lexica}
    \vspace{-5pt}
\end{table}

     

\section{LLMs for Narrative Style Extraction}
Our work explores how LLM-extracted narrative features can be used to yield empirical social insights around empathy and storytelling. As such, we validate whether LLMs are capable of narrative style annotations in line with expert human judgments.
To this end, we prompt GPT-4\footnote{We used gpt-4-0613 accessed via the OpenAI API.} and the instruction-tuned variant of Llama 3 8B\footnote{\href{https://huggingface.co/meta-llama/Meta-Llama-3-8B-Instruct}{meta-llama/Meta-Llama-3-8B-Instruct}} with the same instructions and codebook given to human annotators (Appendix \ref{codebook}). In Table \ref{tab:gptagreement}, we report agreement between averaged human ratings and the LLM-based ratings on the same 50 sampled stories. 

We observe similar patterns in agreement between GPT-4 and human raters as we do in agreement between our two expert annotators. GPT-4 provides ratings with substantial agreement for narrative features such as \textsc{character vulnerability}, 
\textsc{optimistic tone}, and \textsc{resolution}. For most features, the GPT-4 ratings are more positively correlated with human annotations than are the Llama 3 ratings. As such, we use GPT-4 to extract the narrative elements for all the remaining stories in our corpus and  exclude features that have low agreement with human gold labels in our subsequent empirical study. 


\subsection{Performance of LLMs vs. Lexica}
As prior works use lexica \cite{roshanaei_paths_2019, zhou_language_2021} to quantify narrative elements, we compare whether GPT-4 and Llama 3 can outperform psychologically validated lexica in capturing features of \textsc{Heart}. We select 4 dimensions in our taxonomy that readily map to lexicon-based dimensions in LIWC-22 \cite{boyd_development_nodate, pennebaker_linguistic_1999} and compare correlation to human expert ratings in Table \ref{tab:lexica}. We find that GPT-4-extracted features for \textsc{optimistic tone}, \textsc{vivid emotions}, and \textsc{character vulnerability} are better aligned with human ratings than LIWC correspondents, although only \textsc{character vulnerability} is statistically significantly higher ($p<0.001$ as measured by Fisher's exact test). However, LIWC outperforms GPT-4 in the \textsc{Cognition} category, although not statistically significantly so. We discuss the source of potential errors in using GPT-4 to extract \textsc{Cognition} level of narratives in our error analyses below. Notably, although Llama 3 annotations are generally relatively less correlated with human annotations, the Llama 3 extracted features consistently outperform the LIWC correspondents.

\subsection{Error Analysis}
We observe that GPT-4 consistently over-rates the level of \textsc{evaluations} and \textsc{cognition} expressed in a story as compared to human annotators. Through qualitative examples of stories where GPT-4 and human disagreements are large (Appendix \ref{errors}), GPT-4 typically conflates emotional reactions with evaluations, attributions, or desires (e.g. ``\textit{...it really got me thinking about when I first went to College...How excited my parents were for me and scared. And I was both excited and scared...}''). 
For \textsc{cognition} errors, we see that these systematic errors are typically due to GPT-4 conflating recollection with demonstrations of cognition when overall, the story did not contain more internal thinking processes.

Regarding Llama 3, we observe that when human annotators and GPT-4 agree, but Llama 3 disagrees, it tends to assign higher scores to a minority of features (e.g., \textsc{character vulnerability}) while giving lower scores to a majority of features (e.g., \textsc{vivid emotions}, \textsc{vivid setting}). The lower ratings for imagery-related features suggest a lesser adeptness with figurative language.


Ultimately, our validation study demonstrates that LLMs -- in particular, GPT-4 -- can approximate extracting narrative elements relevant to empathy as corroborated by prior work \cite{shen_modeling_2023-1, ziems_can_2024}, but some features are more challenging for the model to identify. We show in the following section that GPT-4 narrative ratings still reveal interesting behavioral insights around narrative empathy, even without perfect agreement.

\section{Human Study for Measuring Empathy}
To demonstrate the empirical use of our taxonomy and how extracted narrative elements can be used to explore behavioral insights around narrative empathy, we conduct a large-scale user study presenting stories to different participants and asking them to rate their empathy towards the story. In this section, we discuss our study participants, the task procedure, and our data collection and measures used.

\subsection{Participants}
We recruited $N=2,624$ participants on Prolific\footnote{\url{https://www.prolific.com/}} to read and rate empathy towards personal stories. An overview of participant demographics is shown in Appendix \ref{demographics}. Participants were balanced by sex, predominantly white, and had high trait empathy on average. 

\subsection{Study Procedure}
Our study procedure was determined exempt by our institution's ethics review board.
At the beginning of the study, participants rated their current emotional state (arousal/valence), before reading a personal story. After reading the story, they were asked to rate their empathy towards the story, and to check which of the narrative elements within our taxonomy based on which elements contributed most to their emotional reaction towards the story. We asked a qualitative, open-ended question asking what aspects of the narrative's style made them relate to the story. 

After this, we asked participants to answer questions related to (1) narrative-reader interaction effects, which encompass reader factors that are tied to the process of reading the narrative (narrative transportation, prior experience with something that happened in the story, and perceived similarity to the narrator, and (2) reader characteristics (age, gender, ethnicity, trait empathy, how often they read for pleasure, fluent languages, and education level). Survey measurements and reasoning for selecting such measurements are detailed in the following section.
All participants were paid \$1 for answering the survey, and participants spent on average $7$ minutes completing the entire task. Each of the 874 stories was rated at least 3 times by independent readers, resulting in 2,624 empathetic reactions to stories in total.

\subsection{Data Collection and Measures}
Our user study aims to capture empathy towards a diverse set of narratives with a diverse set of participants with varying reader characteristics in addition to variables that might moderate the effect of narrative style on empathy. Based on related empirical work exploring factors related to empathy (Figure \ref{fig:interactions}), we designed the following surveys (all surveys are included in Appendix \ref{surveys} for reproducibility). We make our dataset publicly available to open up deeper research in narrative empathy analysis.

\paragraph{Empathy and Narrative Style Preferences} We measure empathy towards the story through the State Empathy Scale \cite{shen_scale_2010}. 
To gauge narrative style preferences, participants check off relevant elements from our taxonomy that they felt contributed to empathy towards the story. In addition, we ask for qualitative free-response feedback on what narrative style elements contributed to empathy towards the story.

\begin{figure}[t]
    \centering
    \includegraphics[width=\linewidth]{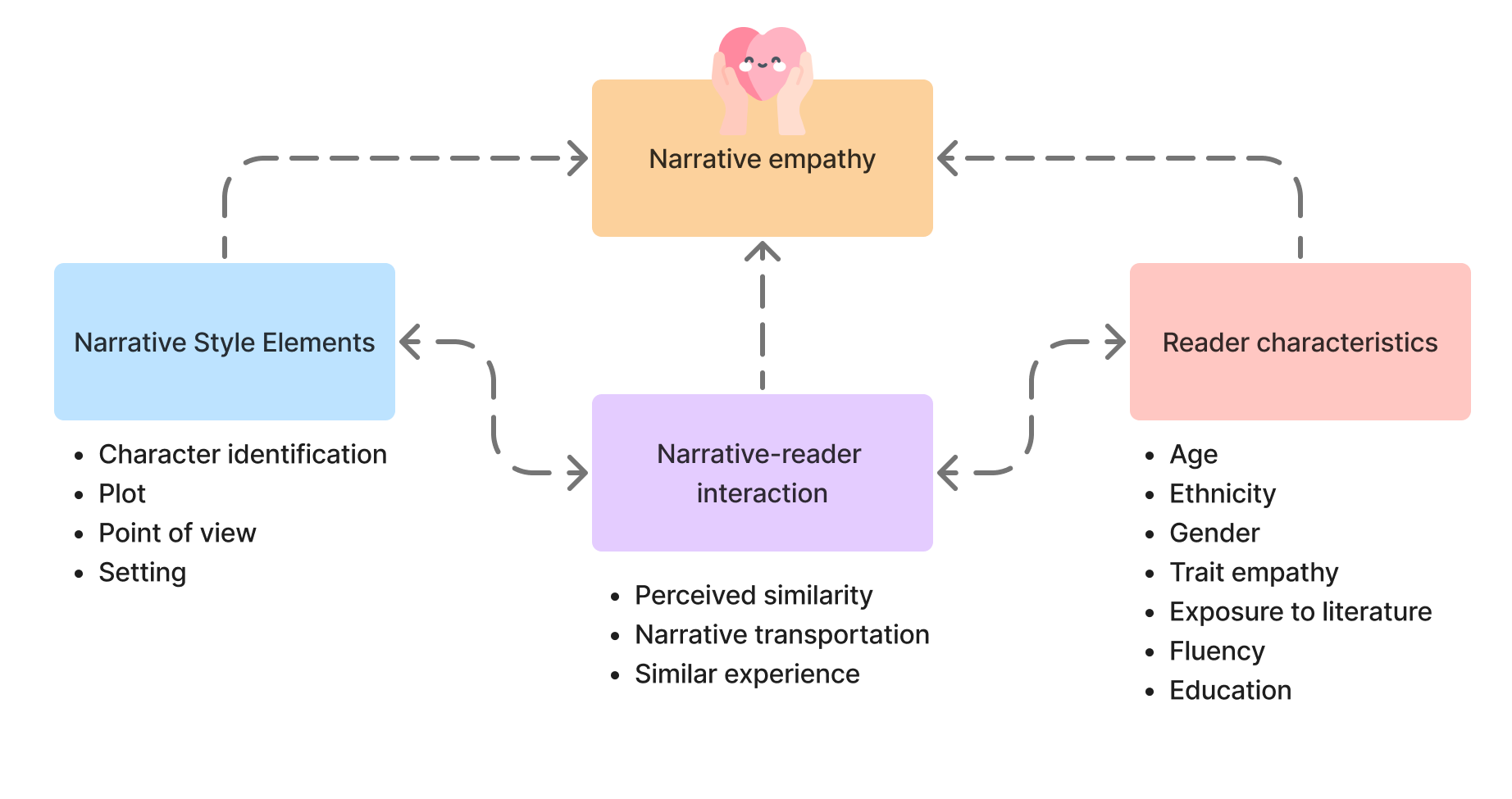}
    \caption{Visualization of how narrative style elements and reader characteristics influence the experience a reader has with a narrative (narrative-reader interaction effects). All of these components combined in turn influence downstream narrative empathy. 
    \vspace{-5pt}
    }
    
    \label{fig:interactions}
\end{figure}

\begin{figure*}[t]
    \centering
    \includegraphics[width=.72\linewidth]{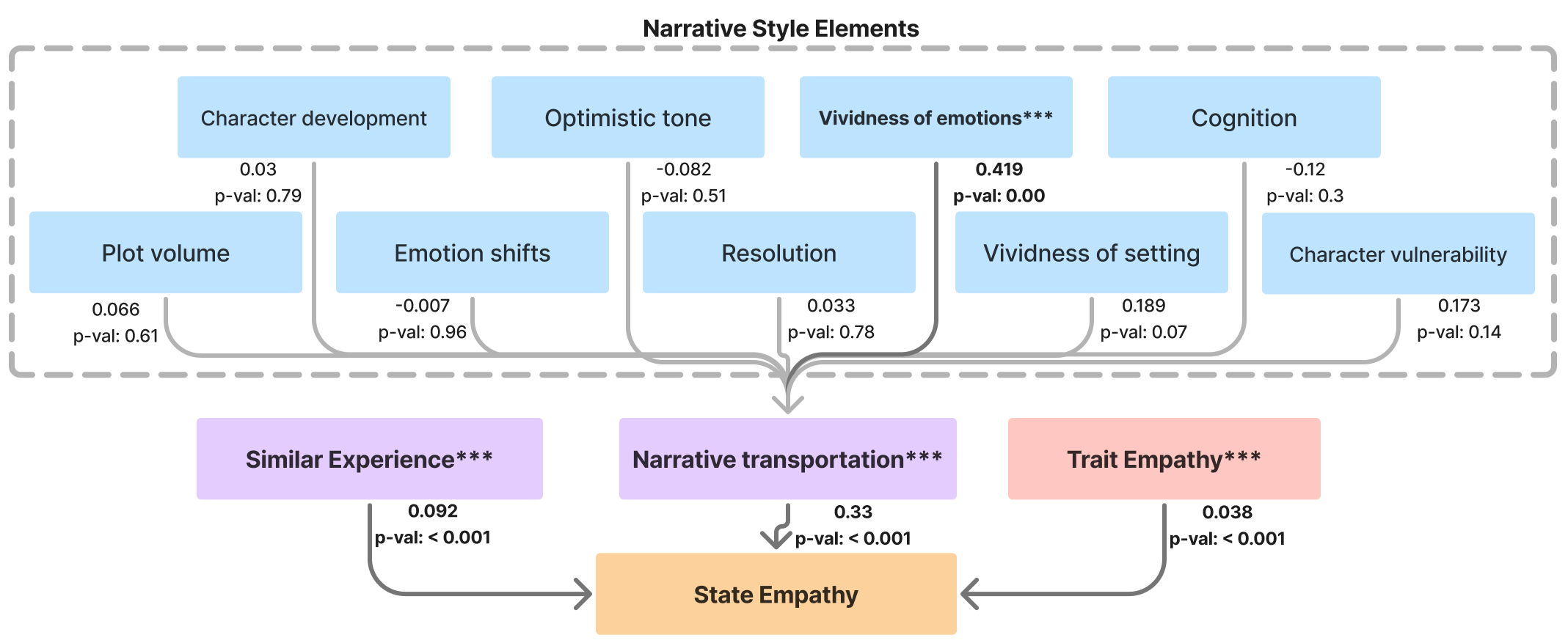}
    \caption{Structural equation modeling of how narrative style elements lead to narrative transportation, combined with effects of the reader sharing a similar experience with the narrator and the reader's baseline trait empathy. }
    \label{fig:SEM}
    \vspace{-5pt}
\end{figure*}

\paragraph{Narrative-Reader Interaction Effects} We define effects at the intersection of reader characteristics and the experience of reading the narrative as narrative-reader interaction effects. These include (1) narrative transportation, measured by the Transportation Scale Short-Form / TS-SF \cite{appel_transportation_2015, walkington_impact_2020}
, (2) prior experience, measured by a Likert scale of how much the reader believes they have been in a similar situation as the narrator, and (3) perceived similarity to the narrator, measured by  the Perceived Relational Diversity Scale \cite{clark_perceived_nodate}. These features allow us to better understand the pathways via how narrative style elements interplay with narrative-reader interactions to lead to downstream empathy. 

\paragraph{Reader Characteristics} We collect reader characteristics based on comprehensive literature review of properties that are related to empathy.
These features include (1) the emotional state of the reader before reading the story, measured by the arousal/valence scale \cite{roshanaei_paths_2019}, (2) basic demographic information including age, gender, ethnicity \cite{christov2014empathy, michalska2013age, o2013empathic}, (3) how often participants read for pleasure \cite{koopman_empathic_2015, mar_bookworms_2006}, and (4) trait empathy, measured by the Single Item Trait Empathy Scale / SITES \cite{konrath_development_2018} and the Toronto Empathy Questionnaire / TEQ \cite{spreng_toronto_2009}. Prolific automatically provides additional demographic information on participants such as fluent languages, nationality, and employment and student status.  


\begin{figure}[t]
    \centering
    \includegraphics[width=\linewidth]{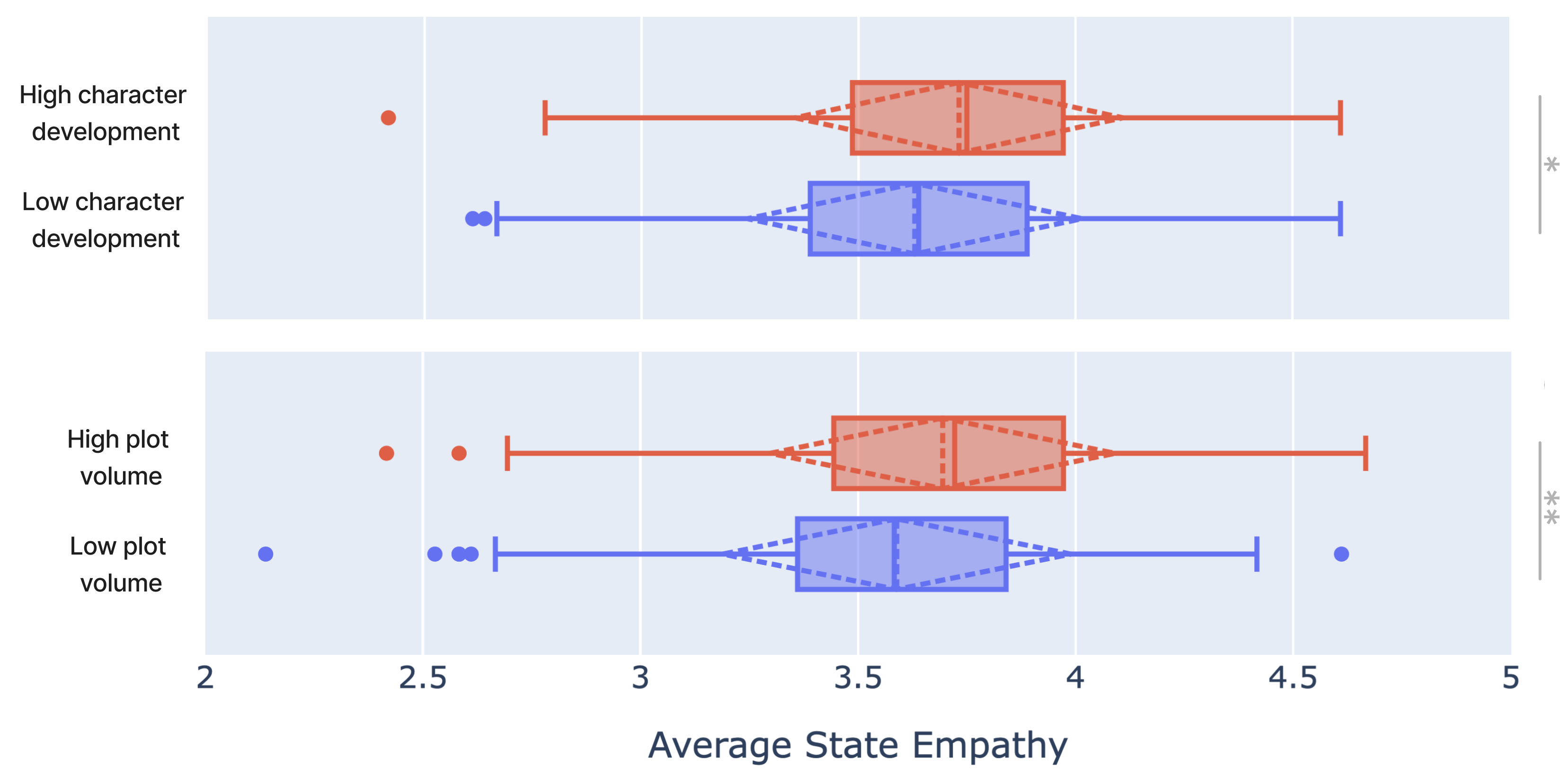}
    \caption{Comparing average empathy across high vs low presence of each narrative feature, we show that there are significant increases in empathy for stories with more character development and plot volume.}
    \label{fig:utests}
        \vspace{-5pt}

\end{figure}
\section{Empirical Insights on Narrative Empathy}
Next, we demonstrate the efficacy of our taxonomy in exploring empirical questions around empathy with a relevant subset of features from our dataset. 

\paragraph{Narrative Style Affects Empathy}
First, we aggregate empathy ratings for each story by taking the mean across the 3 raters. Then, we split stories into high vs. low presence of each narrative feature and apply Mann-Whitney u-tests to the averaged state empathy for the stories. Figure \ref{fig:utests} shows that \textbf{high aggregated empathy stories have more character development and plot volume.} These results are statistically significant, after applying Benjamini-Hochberg correction to account for nine comparisons ($p = 0.03$ for character development, $p = 0.03$ for plot trajectory). 

Our work is, to the best of our knowledge, the first to empirically test the effect of character development and plot volume on narrative empathy. While some prior works \cite{van_krieken_evoking_2017} propose narrative features that relate to character identification, these are lower level than character development, such as the flatness/roundness or vulnerability of a character. Our findings regarding plot volume are in line with prior works that discuss how salient plot events can mark important moments in narratives that influence the emotional impact of the story \cite{sap_quantifying_2022}. Prior works primarily from narrative studies use hand-crafted features on smaller story sets \cite{fernandez-quintanilla_textual_2020, eekhof_engagement_2023}, but do not find significant effects of narrative features such as viewpoint and foregrounding. These studies focus primarily on literary texts rather than narratives that are more common online, and do not take into account other aspects of narrative style and narrative traits that are a part of our theorized taxonomy. These findings suggest future focused works, for example looking at how narrative style relates to empathy across narrative forms (literary vs. personal stories, spoken vs. textual, etc.)

\paragraph{Narrative Empathy is not ``One Size Fits All''} While our previous analysis captures aggregated empathy, different people can have diverse emotional reactions to the same story. In Figure \ref{fig:stds} (Appendix \ref{stds}), we show the standard deviations in state empathy scores for the same story, finding that on average this std. dev. is
significantly greater than zero ($p<0.001$), indicating that the same narrative can evoke different levels of empathy. To address within-subject variance, we fit mixed effects models of empathy ratings using demographic groups, grouping individuals of similar Age, Sex, Trait Empathy, and Ethnicity and conditioning on multiple ratings for a single story. We find through a likelihood ratio test that empathy predicted by demographic group results in significantly better model fit ($p=0.002$). These two results indicate that there is high variance in empathy for the same story and that incorporating information regarding diverse demographic profiles can improve empathy model fit, aligning with prior works \cite{august_exploring_2020}. Our findings have implications in broader empathy prediction tasks within NLP \cite{buechel_modeling_2018-1, sharma_computational_2020}, which often optimize for a single objective empathy score assigned to a piece of text, aggregating empathy which can overlook individual factors.

\paragraph{Vivid Emotional Expression of Narratives Leads to Narrative Empathy}
Given our finding that narrative empathy is not ``one size fits all,'' we conduct analyses taking into account random effects for each story ID with structural equation modeling using the semopy library\footnote{\url{https://semopy.com/}}.
Structural equation modeling (SEM) is a standard social science method for structured hypothesis testing and uses a formulation of generalized linear models to account for fixed and random effects when a theoretical model with relationships between elements is proposed. 

From our SEM results (Figure \ref{fig:SEM}), we find that \textbf{vividness of emotions significantly impacts narrative transportation, which in turn influences downstream empathy towards the story. }The importance of vividness of emotions in personal stories is supported by other work in psychology. In particular, \citet{pillemer_remembering_1992} elaborates that vivid descriptions of emotion in personal stories can convey believability in the experience, more readily evoking empathetic responses. While some computational works explore impact of narrative features on empathy \cite{roshanaei_paths_2019}, they typically focus on positive/negative emotion words, rather than the narrative style or way in which emotions are conveyed through text, and may be better captured by current large-language models.

Figure \ref{fig:SEM} shows how narrative features contribute to narrative transportation, leading to downstream empathy and taking into account non-stylistic factors like the reader sharing a similar experience as the narrator and the reader's trait empathy level. \textbf{We find that both the narrator's previous experience with something happening in the story as well as their baseline trait empathy are significant predictors of empathy towards the story, but not as much as narrative transportation.} In particular, our findings are in line with appraisal theory that suggests that feeling similar emotions is predicated on the target sharing similar experiences \citep{wondra_appraisal_2015, yang_modeling_2024}. While it is not particularly surprising that similar experience correlates with empathy, very few works have looked at narrative style interactions in tandem with fixed (trait empathy) and more dynamic traits (experiencing something similar), suggesting more holistic consideration of contextual factors related to narrative empathy.


\paragraph{Narrative Style Preferences in Relation to Empathy are Personalized}
Finally, we show different demographic profiles might prefer different ways of telling a story, where preference is gauged by narrative empathy. Adding the interaction term  \textsc{Trait Empathy} $\times$ \textsc{Vividness of Emotions} to our structural model, we find a significant interaction effect of vivid emotions on the state empathy (est = $0.252$, $p < 0.001$). This indicates that \textbf{the relationship between vividness of emotions and state empathy increases as trait empathy increases, suggesting that narrative style preferences are personalized across demographic profiles.}

While certainly not exhaustive, our empirical analyses show how \textsc{Heart} can be used to yield interesting behavioral insights around how narrative style contributes to empathy. In particular, we note that looking at \textit{personalization} in narrative empathy, as well as \textit{contextualizing reader factors} such as their trait empathy level are important for empathy prediction, and are often overlooked in existing empathy tasks. 

\section{Conclusion}
In this work, we quantify narrative style as it relates to narrative empathy. We introduce \textsc{Heart}, the first theory-driven taxonomy delineating elements of narrative style that can evoke empathy towards a story. We evaluate the performance of LLMs in extracting narrative elements from \textsc{Heart}, showing that prompting GPT-4 with our taxonomy leads to reasonable, human-level annotations beyond what prior lexicon-based methods can do, but that LLMs struggle in specific tasks, such as GPT-4's limited ability to extract expressions of cognition and evaluations. Through a crowdsourced study with over 2,000 participants, we demonstrate how \textsc{Heart} can be used to empirically understand the empathic role of  narrative style factors. We find that vividness of emotions expressed, character development and plot volume are related to narrative empathy, and contextual factors such as a person's baseline trait empathy or sharing an experience with the narrator contribute to these effects. Additionally, we show that empathy responses are highly variable even in the same story, and that narrative style preferences are personalized to people with different demographic profiles (such as varying levels of trait empathy). 
Our findings show the promise of using LLMs for annotating complex story features that can yield interesting social and behavioral insights. 

\section*{Limitations}
\paragraph{Narrative Style Annotation} While most of the features in our taxonomy yielded reasonable consistency across human and LLM annotators, a few elements such as bodily perception and evaluations were less consistent. We excluded these features from our empirical analysis, but future work could make improvements to the annotation process for these specific elements. For example, our codebook makes use of Likert scale ratings for each of the narrative features within an entire story, but more granular annotations such as frequency of occurrences may have more consistency.

\paragraph{Empirical Study Size and Reproducibility} Findings in human behavior should be reproducible across different populations and contexts. While we conducted a large scale study with many participants, we did not ask participants to rate multiple stories. Additionally, the demographic distribution of Prolific crowdworkers is predominantly white. Future work should aim to reproduce our empirical insights with diverse populations and different types of stories.

\paragraph{Statistical Modeling} 
Our analysis methods involve interpretable statistical models commonly used in social science research. We chose to use structural equation modeling to gauge behavioral insights around how narrative style contributes to empathy, rather than achieving the best performance on narrative empathy prediction. Future work could improve upon narrative empathy prediction by incorporating narrative features in more complex transformer-based models and ablating different features.

\section*{Ethical Considerations}
Personal stories can contain intimate and vulnerable information, in addition to inducing emotions in readers. Our study protocol for showing sensitive stories to crowdworkers was approved by our institution's ethics review board as an exempt study. Participants gave informed consent that their survey ratings would be collected via Prolific. 
We ensured that all datasets we used were also collected via IRB-approved protocols, and will only distribute our dataset to IRB-approved protocols.

More broadly, our work aims to advance research in narrative analysis as it relates to real-world human outcomes, such as empathy. Our findings corroborate that empathy is a highly personalized and contextualized experience. As such, in future work, we find that, rather than modeling the average person, it is important to value the rich diversity of human experiences.

We recognize the ethical implications of modeling empathy in stories is double-edged. Empathy can be used in persuasion,  marketing, or emotional manipulation. We encourage the findings from our work, and future work on narrative empathy analysis, to focus on improving human empathy for social good. For example, one could develop interactive tools to help a user convey a story more empathetically through understanding the role of narrative devices in reader empathy. Or one could use these insights to understand, at scale, social patterns behind storytelling, and how these might drive empathetic shifts online.

\section*{Acknowledgments}
We would like to thank all of our participants and teammates for their invaluable contributions to this project. Special thanks to Sue Holm for narrative annotation and Laura Vianna for study analysis guidance. This work was supported by an NSF GRFP under Grant No. 2141064 and partially funded by NSF grant No. 2230466. 



\bibliography{custom}

\appendix
\section{Participant Demographics} \label{demographics}

\begin{table}[h]
    \centering
    \caption{Participants' demographic breakdown.}
    \resizebox{\linewidth}{!}{
    \begin{tabular}{c|c|c|c|c}
\hline \textbf{Gender} & \textbf{Age} & \textbf{Ethnicity} & \textbf{Trait Empathy} & \textbf{Reading for pleasure}\\
\hline $\begin{array}{l}\text { Female:  1329 } \\
\text { Male:  1295}\end{array}$ & $\begin{array}{l}43  \pm 14 \\
\min : 18 \\
\max : 80\end{array}$
& $\begin{array}{l}
\\
White:  2234 \\
Asian: 150\\
Black: 109\\
Mixed: 86\\
Other: 38 \\
NA: 8\\ 
\\
\end{array}$  & $\begin{array}{l}4.14  \pm 0.88 \\
\min : 1 \\
\max : 5\end{array}$ & $\begin{array}{l}3.45  \pm 1.29\\
\min : 1 \\
\max : 5\end{array}$ \\
\hline
\end{tabular}
}
    \label{tab:demographics}
\end{table}
\section{Distribution of Empathy Standard Deviation} \label{stds}

\begin{figure}[h]
    \centering
    \includegraphics[width=1\linewidth]{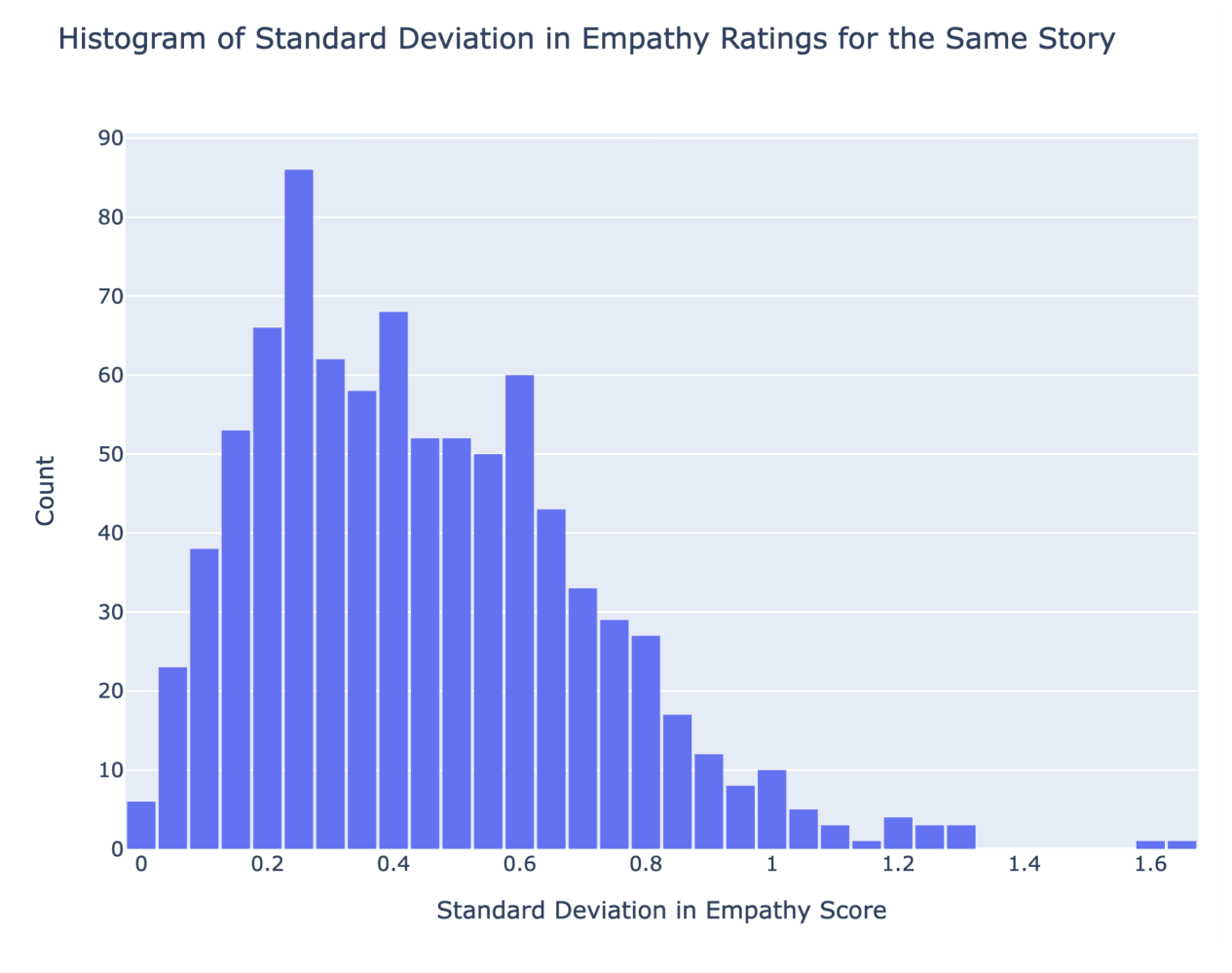}
    \caption{Distribution of standard deviation in empathy scores for the same story indicate that empathy can differ drastically for the same story. }
    \label{fig:stds}
\end{figure}

\section{Codebook and LLM Prompts} \label{codebook}

\subsection{Character development}
We define character development in terms of changes that a character undergoes through the course of narrative events. 

We define changes broadly to include cognitive, emotional, behavioral, spiritual, moral, bodily, and social changes. 

Notably, we do not consider environmental changes for characters sufficient for character development, but acknowledge that other types of change (e.g. emotional, social) often accompany or are caused by environmental changes.

Rate the narrator’s character development based on the following scale:
\begin{itemize}
    \item 1 - no change
    \item 2 - limited change
    \item 3 - moderate change
    \item 4 - significant change
    \item 5 - life-altering, dramatic change
\end{itemize}

Examples:
\begin{itemize}
    \item I watched the birds splashing in the puddle from a bench at the park. They were so playful and content, even as it started to drizzle. (1 - character does not change)
    \item It wasn’t until my brother told me what he’s been going through that I realized how distant I had been. I broke down in front of him at the time. From that day forward, I decided to be there for my family, no matter what–even if that meant quitting my job and moving home. (5 - multiple dramatic changes)
\end{itemize}

Respond with a single integer.

Story: [STORY]

\subsection{Character vulnerability} 
Rate how emotionally vulnerable the narrator is in telling their story. We define vulnerability as how personal or intimate the information shared by the narrator is.

Use the following scale:
\begin{itemize}
    \item 1 - not vulnerable at all
    \item 2 - somewhat vulnerable
    \item 3 - very vulnerable
\end{itemize}

Examples:
\begin{itemize}
    \item But I just doubt myself a lot. It's inevitable. (3 - the author reveals their self doubt)
    \item I went on a very memorable trip to Crater Lake Oregon on July 8th. (1 - does not share any sensitive information)
\end{itemize}

Respond with a single integer.

Story: [STORY]

 \subsection{Optimistic tone}
 Rate the level of optimistic/pessimistic tone in the narrator’s story. This should be the tone from the narrator’s perspective, not of other characters in the story.
 
\begin{itemize}
    \item -2 - very pessimistic
    \item -1 - somewhat pessimistic
    \item 0 - neutral
    \item 1 - somewhat optimistic
    \item 2 - very optimistic
\end{itemize}

Examples:
\begin{itemize}
    \item I feel alone. It is so frustrating. I used to be fine with it, but then for some reason I actually started wanting to have a friend. I have nobody. And nobody around me seems interesting enough to me. Life gets boring. And frustrating. (rating = -2, very pessimistic)
    \item He is grown up and I have done my job to get him out into the world. I will miss his teenage years (somewhat), but I am proud of him. (rating = 2, very optimistic)
\end{itemize}

Respond with a single integer.

Story: [STORY]

\subsection{Vivid emotions}
Rate the vividness of emotions described in the story. For example, vividness can be characterized by metaphor, simile, imagery, or strong language.

Use the following scale:
\begin{itemize}
    \item 1 - not vivid at all
    \item 2 - somewhat vivid
    \item 3 - very vivid
\end{itemize}

Examples:
\begin{itemize}
    \item I didn’t feel great about the situation. (1)
    \item He was a hard-hitter in business, but outside of work he was completely different. (2)
    \item The pain of losing someone is like being stabbed in the chest. I was devasted when I lost her. (3)
    \item I was totally exhausted, tears running down my face (3).
\end{itemize}

Respond with a single integer.

Story: [STORY]

\subsection{Expressions of cognition}
Rate how prominent descriptions of cognitive processes are in the story. We define descriptions of cognitive processes as statements that reveal the mental state or thinking pattern of the narrator.

Use the following scale:
\begin{itemize}
    \item 1 - minimal or no cognitive processes
    \item 2 - moderate prominence of cognitive processes
    \item 3 - high prominence of cognitive processes
\end{itemize}

Examples:
\begin{itemize}
    \item I was born in the United States. (rating = 1)
    \item I wondered if I had seen him before. (rating = 2)
    \item I was thinking about how I could do it but I couldn’t focus because I kept remembering what Sam said to me yesterday. (rating = 3)
\end{itemize}

Respond with a single integer.

Story: [STORY]







\subsection{Temoral references }
Rate the extent to which the character focuses on the past (such as expressing nostalgia or reflections on memories) vs on the future (anticipation, looking forward) in the context of the story. 

Note that we are not asking whether the story is a past-tense, present-tense, or future-tense story. We are concerned with the orientation the narrator has toward the past, present, or future. We define ‘extent’ as the amount of narration time oriented toward the relative past, present, or future. 

Use the following scale:
\begin{itemize}
    \item -2 - heavy focus on the past
    \item -1 - light focus on the past
    \item 0 - focus on the present
    \item 1 - light focus on the future
    \item 2 - heavy focus on the future
\end{itemize}

Examples:
\begin{itemize}
    \item “I was stuck in bureaucratic processes for a year, and the whole time I was dreaming of the day my application processing was complete.” (2)
    \item “I went to the mall and saw a parade on my way.” (0)
    \item “When I started taking my test, I regretted how little I had studied.” (-1)
\end{itemize}

Respond with a single integer

Story: [STORY]

\subsection{Plot volume} Stories are structured by a sequence of events.

We define plot trajectory as the amount and significance of events in the story. 

If the events are banal or insignificant and do not have a big impact on characters, then the plot trajectory is relatively small. If the events significantly impact characters or setting, then the story has a large plot trajectory. 

Rate the degree to which characters and setting are transformed through the course of the story based on the following scale:
\begin{itemize}
    \item 1 - no change
    \item 2 - trivial change
    \item 3 - moderate change
    \item 4 - significant change
    \item 5 - life-altering, dramatic change
\end{itemize}

Examples:
\begin{itemize}
    \item I stared out the window absent-mindedly for three hours. It was a lovely day. (1)
    \item I heard a crash outside. I ran outside to see what had happened. It turned out the wind had blown over a box of garden tools. (3)
    \item After a long and difficult pregnancy, I gave birth to a beautiful baby at 4:15pm. It was a crazy day at the hospital, but thanks to my family and the medical staff, we got through it! (5)
\end{itemize}

Respond with a single integer.

Story: [STORY]

\subsection{Emotion shifts}
Most (but not all) emotions have either a positive (high) or negative (low) valence.  

For example, “anger” and “disgust” are low valence, whereas “happy” and “content” are high valence. 

Other emotions like “ambivalent” or “surprised” could be neutral, low, high, or ambiguous depending on the context.

We consider 5 different types of emotional shifts that can occur in a story:
\begin{itemize}
    \item low-to-high valence (e.g. sad to happy)
    \item high-to-low valence (e.g. happy to sad)
    \item high-to-high valence (e.g. happy to hopeful)
    \item low-to-low valence (e.g. sad to angry)
    \item ambiguous-to-any valence (e.g. bittersweet to excited)
\end{itemize}

We are interested in relatively straightforward emotion shifts that are either explicitly asserted in the text or easily inferrable based on information in the text. We are less interested in extremely subtle emotional shifts (e.g. joyful to content).

Rate the degree of emotional shifts in the story below.

Use the following scale:
\begin{itemize}
    \item 1 - no emotional shifts
    \item 2 - limited or trivial emotional shifts
    \item 3 - moderate emotional shifts
    \item 4 - significant emotional shifts
    \item 5 - life-altering, dramatic emotional shifts
\end{itemize}

Example Story:
\begin{itemize}
    \item “I went to college for 1 year before dropping out.” (1)
    \item “I was surprised to see my friend show up at the cafe where I was working” (2)
    \item “I was frustrated with Ben for not inviting me, but when I ran into him a few weeks later, our conversation went fine.” (3)
    \item “I worked hard all semester and was mentally and physically exhausted by the end. It was such a relief to see my grades come in and see that all of my hard work paid off.” (4)
    \item “I was so excited to get out of class but before the bell rang the principal called me to his office. I was in trouble. I was stressed out of my mind walking to his office, but when I got there, he gave me the good news: I won the school-wide design contest!” (5)
\end{itemize}

Respond with a single integer. 

Story: [STORY] 

\subsection{Resolution}
 In the course of events and interactions between characters, stories introduce conflict. Stories also raise questions about the motives of characters, the meaning of events, and more. Conflict can be explicitly or implicitly referenced by narrators. Alternatively, the reader may subjectively perceive conflict in the situations described by narrators.

Resolution refers to the extent to which conflict is addressed and questions are answered by the end of the story. There are many ways a story may be resolved, partially or completely. Resolution can occur for the narrator, characters within the story, or the reader.

A story with low resolution may not have much conflict or leave conflict unaddressed by the end of the story. A story with high resolution will involve conflict that is addressed or raise questions that are ultimately answered. 

Rate the degree of resolution by the end of the story based on the following scale: 
\begin{itemize}
    \item 1 - no resolution
    \item 2 - limited resolution
    \item 3 - moderate resolution
    \item 4 - significant resolution
    \item 5 - complete resolution
\end{itemize}

Examples:
\begin{itemize}
    \item I couldn’t believe that he didn’t apologize. How can someone just pretend that nothing happened? (1)
    \item I was homeless and finally found a new job, but I hate it and want to find a new one. (3)
    \item I looked for love my entire life, and had almost given up, when I met them. Now I couldn’t be more in love. (5)
\end{itemize}

Respond with a single integer. Do not include any words or punctuation marks in your answer.

Story: [STORY]

\subsection{Vividness of setting}
Rate the vividness of the setting described in the story. For example, vividness can be characterized by metaphor, simile, imagery, or strong language.

Examples:
\begin{itemize}
    \item I went to the restaurant to grab a bite to eat. (1)
    \item The sun cast warm rays onto the concrete in the park. (3)
    \item The waves in Palos Verdes crashed against the shore, making beautiful ribbons (3)
    \item There was a house, with music playing in it (2)
\end{itemize}

\begin{itemize}
    \item 1 - not vivid at all
    \item 2 - somewhat vivid
    \item 3 - very vivid
\end{itemize}

Story: [STORY]

\section{GPT-4 Error Analysis Story Examples}\label{errors}
Scores range from 0 to 1, where 0 indicates low presence of narrative feature and 1 indicates high presence.

\subsection{Stories with \textsc{Evaluations} Disagreement}

Human:  0.0 \\
GPT-4:  1.0 \\

Yeah, so this is the beginning of the school year, and I've seen a lot of people moving into their dorms and apartments, and it really got me thinking about when I first went to College, when I was moving into the dorm. How excited my parents were for me and scared. And I was both excited and scared, moving away from home and my parents, and knowing that I'd probably get really homesick. Watching all those kids moving in really made me think about how that felt for me. It was really important for me, there was a lot of pressure for me to do well in College because my dad came here from a developing country and wasn't able to get an education past second grade. I was the first person in my family to go to College, and to make it that far. So I had a lot of emotions, definitely some anxiety, some stress about the pressure of performing and doing well. But also the excitement, and kind of the normal fear that you get doing something you've never done before and not having parents who had never experienced College like that before. I didn't really have anyone to go to, to understand what that meant, like what to expect. So watching those kids just brought me back to that moment.\\
=================\\
Human:  0.25 \\
GPT-4:  1.0 \\

When covid first hit I had to move from my city to my home town. Wasn't a huge deal as it was for others. It was a year and a half and a lot happened and eventually I came back to my original town, new fiance and cat in tow. I couldn't find a job once I came back and when I did that place got shut down. My fiance did have one, but any paycheck he had didn't go to our shared place, mostly his phone bill or groceries once in a while, while he stayed out till 2 am drinking. So I started. Excessively. 

I don't know now how I did, it was only a few months ago, how I managed to. I borrowed from family or friends, I took out loans to make sure rent was paid. It got so bad I was hospitalized for two weeks because for over a day I was throwing up every hour. I had torn my esophagus. No food. My fiance didn't visit, saying he was scared, but his aunt visited. I cried every night. 

When I got home he was working, he came back and was clearly drinking. The next day my mother texted me telling me she was disappointed in me drinking so much I got myself in that situation. I saw that and realized when I was in the hospital, alone, afraid, and not wanting to live like this. Thus me and my mother didn't speak for months. 

Family stuff happened, got back in touch. We both never apologized but there's a whole lot of cans there to be opened. I ruined relationships I can't take back. Lost contact with friends. 

After this stint my anxiety has been on high alert, making it hard for me to eat or even drink water. Public transit has been scary as I do have those impulse "what if I ran on the track" thoughts which I know are ridiculous. 

Thankfully now I do have a job I enjoy, I have my cat, my own place I pay rent. \\
================= \\
Human:  0.25 \\
GPT-4:  1.0 \\

About four months ago, my wife and I sold our first family home. We have a large family. It is my wife and I plus five children. Our oldest daughter started asking us about having her own room. Although we loved that house, we knew it was time to get something bigger. 

Luckily, we sold it after being on the market for only three days. We found a house with more bedrooms quickly, and the whole process was as smooth as it could have been. 

However, it is bittersweet looking back on everything. That house was very special to me. I did a lot of work on it. I saw my children grow and learn and love there. We made so many good memories. We charted how our children grew on a closet door (which is probably still there). It was a wonderful house while it lasted, but things happen and we had to let it go. 

As of today, I can still remember every little nook and cranny of that place. After all, it has only been a few months. However, it is sad to think these memories will eventually fade. 

I love our new house, but that first one will always hold a special place in my heart.\\
=================\\
Human:  0.0 \\
GPT-4:  1.0 \\

A couple months ago my younger brother got married. I traveled back to my home town of St. Louis, Missouri for the event. I took my girlfriend along with me. It was her first trip to my home town ever. The trip started out great, we picked up our rental car and went to grab a pizza. 

The following day was my brother's wedding ceremony. We thought it had all been planned out thoroughly, but it turns out that nobody had checked the weather report. The day of the wedding came, it started out sunny, a hot day in late July. Clear blue skies and not a cloud to be seen. We were all so optimistic about the big day. The ceremony was scheduled for 6 PM at sunset, so it would start to cool down and allow for some reprieve from the heat of the day. In theory, that was a great idea. 

However, when my girlfriend and I pulled out of the driveway we noticed something new that we hadn't seen yet on the trip. Storm clouds moving in fast, and lots of them. Dark gray giants rose onto the horizon at a frightening pace. Lightning was visible in the distance as we began our drive to the wedding venue. We hoped and prayed that the storm would blow the other way, and that the outdoor wedding venue would be spared from this particular storm. Would we be able to get away with having the ceremony in decent weather? 

It became a race with time. As we drove to the wedding ceremony, it felt as though the clouds were following us and growing larger. As we arrived, I greeted my brother in the parking lot and asked if he thought it would rain. He said maybe, it depends how fast we can get this done. Everyone was present except for the minister, one of the few people who was completely essential to the process. 

As the minister arrived, it finally began to rain. It was raining on my brother's wedding day, I couldn't believe it, but luckily the ceremony was completed and we had a wonderful sunny reception the next day.\\
=================\\
Human:  0.25 \\
GPT-4:  1.0 \\

I was hiking near Lake Ontario with my partner and our two grandkids. It was a beautiful sunny day. The lake sparkled brilliantly. I have a bad knee, so I was struggling with some of the physical activity. My partner suggested that I rest a bit on a fallen log. I was nervous, because I would not be able to get up by myself, but I agreed. 

My partner and my granddaugther wanted to hike further to see the bluffs. I said I would be OK for a bit, but my sweet grandson insisted on staying with me. He said, "I won't let my granny sit in the forest all alone." 

Well it was a good thing. My partner and granddaughter didn't return in a reasonable amount of time. We got very nervous! My grandson is 10 years old, but not strong enough to help me up. He searched for a stout walking stick and found one nearby. I used it to prop myself up, and managed to get my feet underneath me. 

Together we went down the shore to find the rest of our party. Luckily all was well, but I would truly have been distraught if I had been all alone waiting for so long!

\subsection{Stories with \textsc{Cognition} Disagreement}
Human:  0.25 \\
GPT-4:  1.0 \\

I've been hearing a lot of people saying that MIT students aren't successful as Stanford or Harvard students because there aren't as many well-known MIT CEOs. It seems rather unfair of them to say that because MIT students have contributed a lot to this world from nobel-prize winning theorems to groundbreaking algorithms. 

Also there are lot of MIT grads like David Siegel who went on to found great companies that don't necessarily have a face to the brand like Jobs' Apple or Zuckerberg's Facebook. And on top of that, there many of MIT grads who go on to be CTOs or other types of product managers (sorry for the emphasis on course 6), and without them, the companies would not be the same. 

Above all, out of the "top" institutions, MIT does the most to help lower-income students attain social and economic mobility (I remember reading an article, but can't find the link). 

This is not to say that MIT doesn't have problems, but at the end of the day, I wish people didn't equate fame/status with success. You don't have to be a famous CEO or a CEO in general to be successful. And I'm sure a lot of the people I mentioned didn't end up becoming crazy famous because they value privacy, which is fine! 

And a lot of alumns end up doing what they're interested in regardless of status, which is amazing and also indicative of success! Success can look different for people.\\
=================\\
Human:  0.25\\
GPT-4:  1.0\\

When covid first hit I had to move from my city to my home town. Wasn't a huge deal as it was for others. It was a year and a half and a lot happened and eventually I came back to my original town, new fiance and cat in tow. I couldn't find a job once I came back and when I did that place got shut down. My fiance did have one, but any paycheck he had didn't go to our shared place, mostly his phone bill or groceries once in a while, while he stayed out till 2 am drinking. So I started. Excessively. 

I don't know now how I did, it was only a few months ago, how I managed to. I borrowed from family or friends, I took out loans to make sure rent was paid. It got so bad I was hospitalized for two weeks because for over a day I was throwing up every hour. I had torn my esophagus. No food. My fiance didn't visit, saying he was scared, but his aunt visited. I cried every night. 

When I got home he was working, he came back and was clearly drinking. The next day my mother texted me telling me she was disappointed in me drinking so much I got myself in that situation. I saw that and realized when I was in the hospital, alone, afraid, and not wanting to live like this. Thus me and my mother didn't speak for months. 

Family stuff happened, got back in touch. We both never apologized but there's a whole lot of cans there to be opened. I ruined relationships I can't take back. Lost contact with friends. 

After this stint my anxiety has been on high alert, making it hard for me to eat or even drink water. Public transit has been scary as I do have those impulse "what if I ran on the track" thoughts which I know are ridiculous. 

Thankfully now I do have a job I enjoy, I have my cat, my own place I pay rent.\\
=================\\
Human:  0.25\\
GPT-4:  1.0\\

Today was one of the saddest days of my life. It started early in the day, and my parents came by and picked me up at my house. Everyone was in a very somber mood, but it was sunny and quite warm. We drove out to a church about thirty minutes away near where my mom grew up, and while driving I couldn't help but think back to all the good memories I had with my cousin. She was always so happy and nice and just fun to be around. But now, that was all gone, and all I had were the memories that were going over in my mind. 

Arriving at the church and seeing all of my family, it was hard. It was just so sad, all of it. Seeing my aunt was the hardest part I think, but I knew then that she was strong and was going to be able to get past this. My uncle is an ordained pastor, so he was able to help with the service and I think that helped ease some of the pain. 

After the service we all went to the cemetery and gathered up on the hill in the shade. Seeing the final resting place really hit me hard, I started to cry much harder than I had been all day at that point. All of the memories and the final shock to my brain that she was never coming back, made me very sad, and made me miss her dearly. 

We then all met at a local place where they served a late lunch and we had some drinks. It was good to see so many of my family, but at the same time, so sad, because I thought that we shouldn't be seeing each other, at least not for this reason. 

I didn't really know how to feel when we left and I made it back home. I was deeply saddened, and just thought of how my aunt, uncle, and cousins felt. I know that life had changed for them forever, and now life was starting again without their dear one, and that hurt me again. But my family is strong, and stronger together, and I know we will get through this like we will any other tragedy that comes our way.


\section{Surveys}\label{surveys}
\subsection{Empathy and Narrative Style Preferences}
\textit{State Empathy Scale \cite{shen_scale_2010}}

Please indicate the level to which you agree with each of the following statements -- Strongly disagree (1) to Strongly agree (5)

\begin{enumerate}
    \item The narrator's emotions are genuine.
    \item I experienced the same emotions as the narrator while reading this story.
    \item I was in a similar emotional state as the narrator when reading this story.
    \item I can feel the narrator's emotions.
    \item I can see the narrator's point of view.
    \item I recognize the narrator's situation.
    \item I can understand what the narrator was going through in the story.
    \item The narrator's reactions to the situation are understandable.
    \item When reading the story, I was fully absorbed.
    \item I can relate to what the narrator was going through in the story.
    \item I can identify with the situation described in the story.
    \item I can identify with the narrator in the story.
\end{enumerate}

\noindent\textit{Narrative Style Preferences}

Check the aspects of narrative style (the way the story was told) that made you resonate with the story.
\begin{enumerate}
    \item Flatness/roundness of the character (the character shows development/is vulnerable/subverts expectations)
    \item References to the past (nostalgia) or the future
    \item The vividness of emotions described in the story
    \item The way thoughts/cognition of the character are expressed
    \item The way moral judgments of the character are expressed
    \item The way the character's perception and physical sensations are expressed
    \item The way the character's actions are expressed
    \item The way the setting of the story is expressed
    \item The way the character's point of view is expressed
    \item The overall plot trajectory of the story
    \item The presence of a resolution in the story
    \item The flow and readibility of the story
    \item The overall shifts in the emotional tone of the story
\end{enumerate}

[FREE RESPONSE] What about the narrative style (the way the story was told) made you resonate with it (if any)?

\subsection{Narrative-Reader Interaction}

\textit{Transportation Scale Short-Form / TS-SF} \cite{appel_transportation_2015}

Rate the extent to which you agree with the following statements -- Not at all (1) to Very much (7)
\begin{enumerate}
    \item I could picture myself in the scene of the events described in the narrative.
    \item I was mentally involved in the narrative while reading it.
    \item I wanted to learn how the narrative ended.
    \item The narrative affected me emotionally.
    \item While reading the narrative I had a vivid image of the narrator.
\end{enumerate}

\noindent\textit{Similar Experience}

I have experienced a similar situation as the narrator in my life before -- Strongly disagree (1) to Strongly agree (5)

\noindent\textit{Similar to Narrator \cite{clark_perceived_nodate}}
I am similar to the narrator -- Strongly disagree (1) to Strongly agree (5)

Rate how similar you believe you are to the narrator in terms of the following characteristics -- Not similar at all (1) to Highly similar (5)
\begin{enumerate}
    \item Age
    \item Race/ethnicity
    \item Sex
    \item Religion
    \item Sexual orientation
    \item Socio-economic status
    \item Geographic origin
\end{enumerate}

\subsection{Reader Characteristics}
\textit{Education Level}

What is the highest level of education you have completed?
\begin{enumerate}
    \item Some high school or less
    \item High school diploma or GED
    \item Some college, but no degree
    \item Associates or technical degree
    \item Bachelor's degree
    \item Graduate or professional degree (MA, MS, PhD, JD, MD, DDS etc.)
    \item Prefer not to say
\end{enumerate}
\textit{Reading for pleasure}

How often do you read for pleasure?
\begin{enumerate}
    \item Almost never
    \item A couple of times a year
    \item A couple of times a month
    \item At least once a week
    \item Once or more a day
\end{enumerate}

\noindent\textit{Trait Empathy \cite{konrath_development_2018, spreng_toronto_2009}}

To what extent does the following statement describe you: "I am an empathetic person" -- Strong disagree (1) to Strongly agree (5)

Below is a list of statements. Please read each statement carefully and rate how frequently you feel or act in the manner described -- Never (1) to Always (5)
\begin{enumerate}
    \item When someone else is feeling excited, I tend to get excited too
    \item Other people’s misfortunes do not disturb me a great deal
    \item It upsets me to see someone being treated disrespectfully
    \item I remain unaffected when someone close to me is happy
    \item I enjoy making other people feel better
    \item I have tender, concerned feelings for people less fortunate than me
    \item When a friend starts to talk about his/her problems, I try to steer the conversation towards something else
    \item I can tell when others are sad even when they do not say anything
    \item I find that I am “in tune” with other people’s moods
    \item I do not feel sympathy for people who cause their own serious illnesses
    \item I become irritated when someone cries
    \item I am not really interested in how other people feel
    \item I get a strong urge to help when I see someone who is upset
    \item When I see someone being treated unfairly, I do not feel very much pity for them
    \item I find it silly for people to cry out of happiness
    \item When I see someone being taken advantage of, I feel kind of protective towards him
\end{enumerate}


\end{document}